\pdfoutput=1

\documentclass[11pt]{article}

\usepackage[]{ACL2023}

\usepackage{times}
\usepackage{latexsym}

\usepackage[T1]{fontenc}

\usepackage[utf8]{inputenc}

\usepackage{microtype}

\usepackage{inconsolata}
\usepackage{subfig}
\usepackage{multirow}
\usepackage{graphicx}
\usepackage{amsmath}
\usepackage{amssymb}
\usepackage{booktabs} 
\title{Instruction Tuning with Retrieval-based Examples Ranking for \\Aspect-based Sentiment Analysis}

\author{Guangmin Zheng${^1}$, Jin Wang\thanks{\enspace Corresponding author.}${\;\,^1}$, Liang-Chih Yu${^*}$${^2}$ and Xuejie Zhang${^1}$ \\
  ${^1}$School of Information Science and Engineering, Yunnan University, Yunnan, P.R. China\\
  ${^2}$Department of Information Management, Yuan Ze University, Taiwan \\
  \texttt{Contact: wangjin@ynu.edu.cn, lcyu@saturn.yzu.edu.tw} \\}

\begin{document}
\maketitle

\begin{abstract}
Aspect-based sentiment analysis (ABSA) identifies sentiment information related to specific aspects and provides deeper market insights to businesses and organizations. With the emergence of large language models (LMs), recent studies have proposed using fixed examples for instruction tuning to reformulate ABSA as a generation task. However, the performance is sensitive to the selection of in-context examples; several retrieval methods are based on surface similarity and are independent of the LM generative objective. This study proposes an instruction learning method with retrieval-based example ranking for ABSA tasks. For each target sample, an LM was applied as a scorer to estimate the likelihood of the output given the input and a candidate example as the prompt, and training examples were labeled as positive or negative by ranking the scores. An alternating training schema is proposed to train both the retriever and LM. Instructional prompts can be constructed using high-quality examples. The LM is used for both scoring and inference, improving the generation efficiency without incurring additional computational costs or training difficulties. Extensive experiments on three ABSA subtasks verified the effectiveness of the proposed method, demonstrating its superiority over various strong baseline models. Code and data are released at \url{https://github.com/zgMin/IT-RER-ABSA}.
\end{abstract}

\section{Introduction}
Aspect-based sentiment analysis (ABSA) \cite{Zhang2017} is a fine-grained text sentiment analysis technique that identifies sentiment information related to specific aspects, and provides deeper market insights for businesses and organizations. ABSA consists of three subtasks: aspect term extraction (ATE), aspect term sentiment classification (ATSC), and aspect sentiment pair extraction (ASPE). ATE identifies the aspects mentioned in the text, and ATSC determines the sentiment polarity associated with each aspect. Furthermore, ASPE jointly performs ATE and ASPE to extract sentiment tuples, including aspect terms and their associated sentiments.

For ABSA tasks, most previous methods have used transformer-based language models (LMs) in either a pipeline or end-to-end framework \cite{Chen2020,Luo2020,Mao2021,Ricardo,Yang2021}. By adding task-specific layers to the top of the model, these models are typically initialized from the pretrained checkpoint and then finetuned on the downstream samples. Recently, generative models \cite{Hosseini,Yan2021,Zhang2021} have emerged to reformulate ABSA tasks as generation tasks to produce a sequence with a special pattern, for example, \texttt{restaurant\#positive} and \texttt{food\#negative}. Based on the instruction-tuning paradigm \cite{Mishra2022}, several studies \cite{Scaria2023,Varia2022} have further improved the generative approach using several predefined instruction prompts \cite{Scaria2023}. Instruction tuning allows generative models to tune themselves on a few input-output examples.

The quality of the output generated by the instruction-tuning model is highly dependent on the quality of in-context examples. Well-crafted instructions can help the model generate more accurate and relevant outputs\cite{luo2024duetsim}, whereas poorly crafted instructions can lead to incoherent or irrelevant results. Nevertheless, previous studies typically adopted a fixed strategy to use two or more unchanged examples to generate the instruction template. If the examples are unrepresentative of the target task, the model may be unable to learn effectively. 

For a target sample, for example, \textit{The \underline{falafel} was slightly \textbf{overcooked} and \textbf{dry}, but the \underline{chicken} was \textbf{satisfactory}}, the example \textit{The \underline{price} was too \textbf{high}, but the \underline{cab} was \textbf{amazing}} can be appropriate. They share a similar syntactic structure, which can contribute to imitation and generation. However, such an example is unsuitable for another sample, for example, \textit{The \underline{staff} displays \textbf{arrogance}, and the \underline{prices} are considerably \textbf{high} for Brooklyn standards}. Because the opinion of \textit{\textbf{high}} in the example may finally impact the judgment of the aspect \textit{price} of the target. Furthermore, the sample \textit{We \textbf{enjoyed} our \underline{visit} and utilized buses and cabs for transportation} seems to have little relevance to the example above. However, the aspect \textit{cab} may be incorrectly considered an aspect term based on the prompt of the word \textit{cab} in the example. 

Empirical studies have demonstrated the use of an additional language model as a scorer to produce off-the-shelf sentence embeddings, for example, BM25 \cite{Robertson2009}, EPR \cite{Rubin2021} and LLM-R \cite{Wang2023}, for similarity calculations to retrieve examples from the training set. Several works explored training a prompt retriever to select examples by measuring surface similarity \cite{Li2022,Liu2022,Zhang2022}. These methods have two limitations: (i) the similarity calculation typically measures the distance in the latent space, which is independent of the generation target in instruction learning, and (ii) additional encoders are necessary to obtain the representation used for similarity calculations, which also incurs additional computational costs and training difficulties.

This study proposes an instruction-tuning method with retrieval-based example ranking for ABSA tasks, comprising a retriever and inference LM. The LM is a T5 model  \cite{Raffel2019} with an encoder-decoder structure and several instruction-tuned versions \cite{Chung2022,Wang2022}. The retriever returns the most appropriate examples to form the instruction template. Meanwhile, the LM scores the examples for the retriever and generates the final results. 

To achieve the training of LM and retriever simultaneously, an alternating training schema is proposed. For each target sample, the candidates can be divided into positive and negative examples according to the log-likelihood provided by the LM. Then, contrastive learning is applied to force the sample to be near positive examples but distant from negative examples. After composing high-quality instructions, the LM can be finetuned using a generative objective.

Unlike the previously proposed similarity calculation, the retriever evaluates the importance of candidate examples based on the log-likelihood of the LM. This retrieval goal is consistent with the generative objective of the LM. In addition, the LM is used for both scoring and inference, improving the generation performance with tolerable additional computational costs or training difficulties.

Extensive experiments were conducted on ATE, ATSC, and ASPE tasks to verify the effectiveness of the proposed method. The results show that the proposed model substantially improves the performance compared with several strong baselines. 

The remainder of this paper is organized as follows. Section 2 briefly reviews the related studies. Section 3 describes the proposed retrieval-based example mining method for instruction learning in ABSA. Section 4 summarizes the experiment settings and empirical results. Finally, Section 5 concludes the paper.

\begin{figure*}[!t]
    \centering  
    \centerline{\includegraphics[width=1\textwidth]{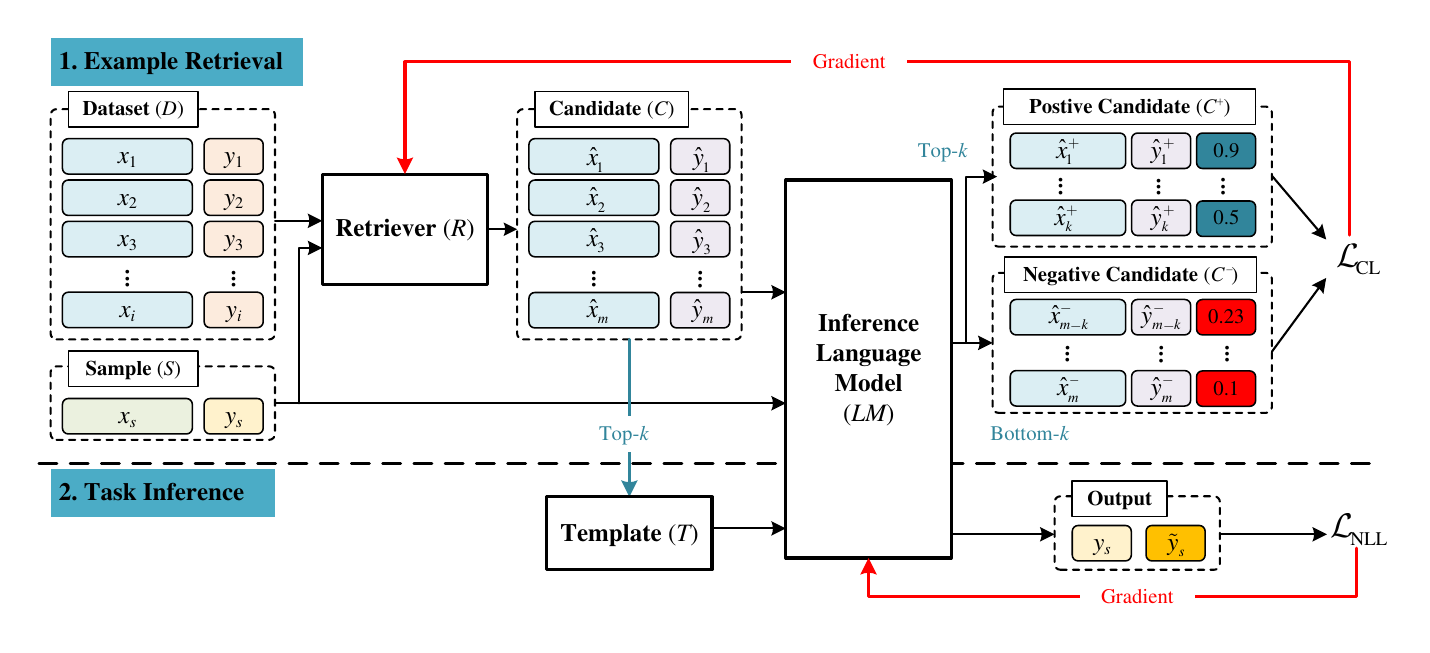}}
    \caption{Overall architecture of the proposed instruction tuning with retrieval-based examples ranking for ABSA.}
    \label{a1}
\end{figure*}

\section{Related Work}
\subsection{Aspect-Based Sentiment Analysis}
Aspect-based sentiment analysis \cite{Zhang2017} aims to analyze sentiment polarities toward the specific aspects of a given text. In ABSA, text is decomposed into several aspects, and sentiment polarity, which is typically positive, neutral, or negative, is analyzed for each aspect. This approach provides fine-grained, in-depth sentiment insights, enabling businesses and organizations to better understand market and consumer perspectives.

Most approaches have focused on using encoder structures to accomplish aspect extraction and sentiment identification, such as improved text encoding using attention mechanisms \cite{Ricardo,yuan2020graph,yuan2022syntactic}, multitask learning \cite{Chen2020}, and approaches based on machine reading comprehension \cite{Mao2021}. Some studies \cite{Yan2021,Zhang2021} introduced decoders to unify ABSA's previous extraction and classification tasks into generative tasks. Recently, instruction prompts have been introduced into generative methods \cite{Varia2022}, achieve a solid few-shot performance by making the model perform the correct action with an apparent task description. Furthermore, fixed examples were added to the instructions to supplement the task description with more accurate information, yielding significant performance improvements in the ABSA subtasks \cite{Scaria2023}. 

However, the adaptability of the fixed examples to different review texts is difficult to guarantee. \citet{Liu2022} demonstrate that selecting contextual examples significantly affects the downstream performance. The intricate relationship between selected examples and diverse review contexts underscores the need for a nuanced approach to ensure robust and effective adaptation in various scenarios.

\subsection{Prompt Retrieval}
With the development of deep learning techniques, dense retrieval \cite{Karpukhin2020} has become a widely used information retrieval method that utilizes dense vectors to semantically match queries and documents in the latent space. Compared with sparse retrieval methods, dense retrieval exploits the powerful modeling capabilities of pretrained language models (PLMs) and may overcome the linguistic mismatch problem. Therefore, dense retrieval has become popular in current retrieval technology.

In context learning, retrieval enhancement aims to improve the performance of LMs in downstream tasks by retrieving information-rich examples \cite{Li2023,Luo2023}. In previous studies, unsupervised sentence encoders have often encoded training examples and retrieved their nearest neighbors for each test instance \cite{Liu2022}. In some studies \cite{Das2021}, a supervised prompt retriever was trained to answer questions on a knowledge base. This retriever relies on the surface similarity to perform queries when it receives supervised training tailored to knowledge-based queries. However, these methods tap only into text associations, ignoring how the language model understands these texts. 

To address this problem, some researchers \cite{Rubin2021,Wang2023} have proposed using the LM to score examples to train retrievers. Unfortunately, these methods are limited because they are only applicable to frozen large language models. This constraint underscores the need for more versatile methodologies that can extend beyond the confines of frozen LMs to enhance the flexibility and generalizability of retriever training strategies.

\section{Retrieval-Based Instruction Tuning}

Figure \ref{a1} shows the overall framework of the proposed instruction-tuning method with retrieval-based example ranking for ABSA. The training process comprises two phases: example retrieval and task inference. For example retrieval, a retriever is used to select several candidates for a given sample. The inference LM was then used to measure the likelihood of ranking the target and candidates as scores. To train the retriever, we selected the top-$k$ candidates as positive samples and the others as negative samples. Subsequently, contrastive learning is performed to propagate the gradients and update the retriever. For task inference, the retriever returns the most suitable examples and forms input instructions according to a predefined template. The negative log-likelihood of the generative results will train the LM.

\subsection{Instruction Template}

\begin{figure}[!t]
    \centering  
    \centerline{\includegraphics[width=0.4\textwidth]{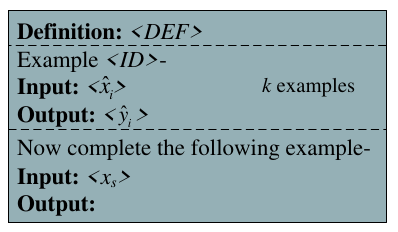}}
    \caption{Template of instruction prompts. \textless$DEF$\textgreater\: is the definition of the task; \textless $ID$\textgreater , the identity of the examples; \textless $\hat{x_i}$\textgreater\: and \textless $\hat{y_i} $\textgreater , the input and output of the examples, respectively; $k$ , the total number of examples; and \textless $x_s$\textgreater , the input text.}
    \label{a2}
\end{figure}

\begin{figure*}[!t]    
  \centering            
  \subfloat[ATE]   
  {
        \label{ea}
        \includegraphics[width=0.325\textwidth]{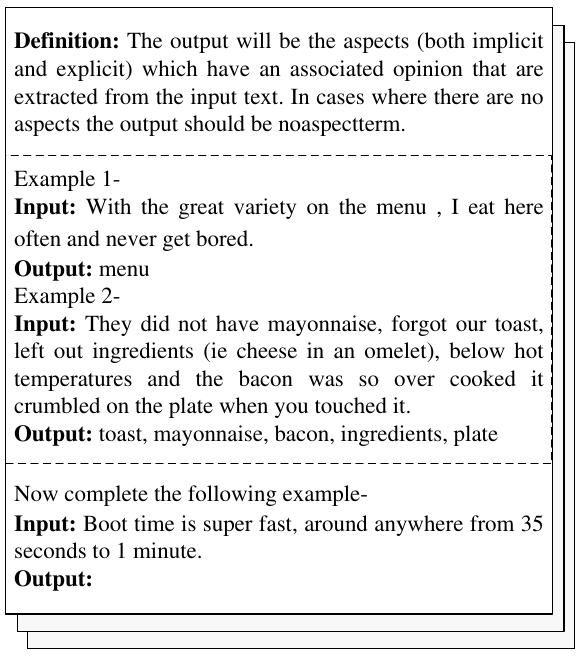}
  }
  \subfloat[ATSC]
  {
      \label{eb}
        \includegraphics[width=0.325\textwidth]{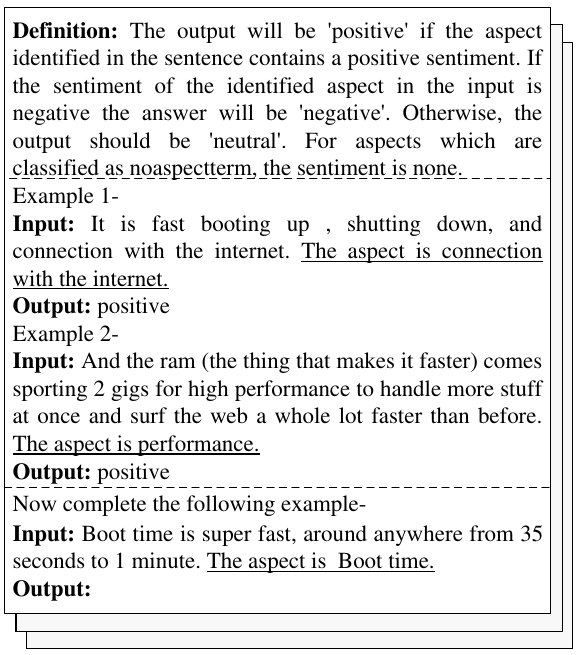}
  }
  \subfloat[ASPE]
  {
      \label{ec}
        \includegraphics[width=0.325\textwidth]{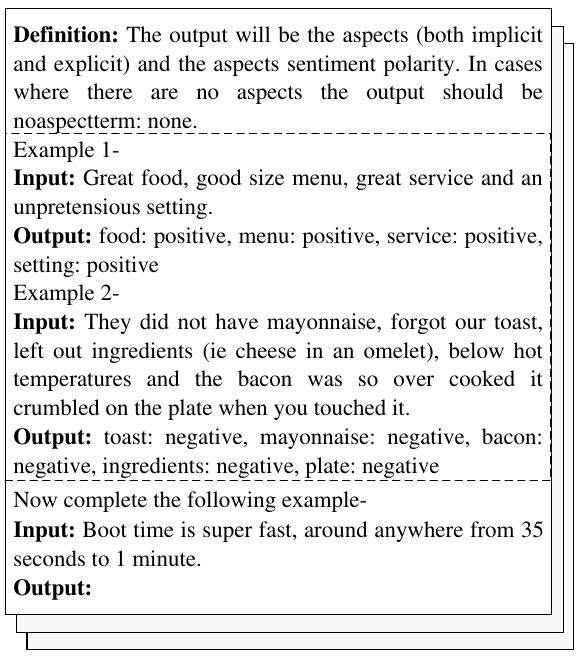}
  }
  \caption{Demonstrations of instruction prompts.}    
  \label{a3}            
\end{figure*}

Figure \ref{a2} shows the instruction prompt template, which comprises a task definition, examples, and input text. In particular, a task \textless $DEF$\textgreater\:is available for different subtasks. For a given input text \textless $x_s$\textgreater, examples \textless $E$\textgreater\:$=\{({{\hat{x}}_{i}},{{\hat{y}}_{i}})\}_{i=1}^{k}$ are selected for instruction tuning, and the output conforms to the formats of the different subtasks. Thus, the instruction template for $x_s$ can be formally denoted as $\text{Tmpl}(DEF,\{{{\hat{x}}_{i}},{{\hat{y}}_{i}}\}_{i=1}^{k},{{x}_{s}})$.

Figure \ref{a3} shows the instruction prompts generated using the ATE, ATSC, and ASPE templates. For ATSC, the aspect term is spliced together with the review text as input through the fixed prompt \textit{The aspect is}, as shown in the underlined portion of Figure \ref{a3}.

\subsection{Example Retrieval}


\noindent\textbf{Candidate Generation.} Training set $D$ consists of several input-output pairs, i.e. $D=\{({{x}_{i}},{{y}_{i}})\}_{i=1}^{n}$, where $x$ is the text, and $y$ is the label. For a target sample $({{x}_{s}},{{y}_{s}})$, retriever $R$ returns the top-$m$ candidates from the training set,
\begin{align}
C=R(({{x}_{s}},{{y}_{s}}),D)=\{({{\hat{x}}_{j}},{{\hat{y}}_{j}})\}_{j=1}^{m}
\label{eq1} 
\end{align}

\noindent Sample $({{x}_{s}},{{y}_{s}})$ should be excluded from the retrieval results.

The LM adopted the encoder-decoder architecture of T5, which scores each candidate independently. The scoring function is the log-likelihood of output $y_s$, which is consistent with the autoregressive decoding objective of the LM, and is denoted as
\begin{align}
  & {{\Delta }_{i}}=\log p({{y}_{s}}|DEF,{{{\hat{x}}}_{i}},{{{\hat{y}}}_{i}},{{x}_{s}}) \nonumber \\
 & \ \ \ \ =\sum\limits_{l=1}^{L}{\log }p(y_{s}^{l}\mid DEF,{{{\hat{x}}}_{i}},{{{\hat{y}}}_{i}},{{x}_{s}},y_{s}^{<l})
 \label{eq2}
\end{align}

\noindent where $p({{y}_{s}}|DEF,{{\hat{x}}_{i}},{{\hat{y}}_{i}},{{x}_{s}})$ is the conditional probability of $y$ given task definition \textless $DEF$\textgreater, input $x_s$, and $i$-th candidate $({{\hat{x}}_{i}},{{\hat{y}}_{i}})$.

The candidate set is sorted in descending order according to the score ${{\Delta }_{i}}$. The top-$k$ candidates form the set of positive samples ${{C}^{+}}$, whereas the bottom-$k$ candidates form the set of negative samples ${{C}^{-}}$. Contrastive learning is then applied to train the retriever, allowing the retrieved results to be scored as high as possible.

To reduce the computational cost of the scoring function, only a portion of the training set with ratio $r$ is extracted to select the positive and negative examples for the training of the retriever. 
   
\noindent\textbf{Retriever Training.} A contrastive learning objective trains the retriever. The input is (Input: \textless ${{\hat{x}}_{i}}$ \textgreater\: Output: \textless ${{\hat{y}}_{i}}$\textgreater) for the candidates $c_i=({{\hat{x}}_{i}},{{\hat{y}}_{i}})$ , and (Input: \textless ${{x}_{s}}$\textgreater) for the target sample. Besides, we followed Sentence-T5 \cite{Ni2022} to average the encoder output as the representation for the candidates and target sample,
\begin{align}
{{\mathbf{h}}_{i}}&=\text{Mean}(\text{Enc}({{\hat{x}}_{i}},{{\hat{y}}_{i}})) \label{eq3}\\
{{\mathbf{h}}_{s}}&=\text{Mean}(\text{Enc}({{x}_{s}}))
 \label{eq4}
\end{align}

\noindent where ${{\mathbf{h}}_{i}}$ and ${{\mathbf{h}}_{s}}$ are the hidden representations of the $i$-th candidate and target sample, respectively. The distance for contrastive learning measures their inner products, and is denoted as
\begin{align}
\text{Sim}({{x}_{s}},{{c}_{i}})=\mathbf{h}_{s}^{\top }{{\mathbf{h}}_{i}}
 \label{eq5}
\end{align}

For the target sample, one positive example $c_i^+$ is randomly selected from the set of positive samples $C^+$, one negative example from the set of negative samples $C^-$, and the other negative examples come from $B-1$ positive and $B-1$ negative examples sampled for the other samples in the same batch, where $B$ is the batch size. The objective function minimizes the negative log-likelihood of the positive example for the target sample.

\begin{align}
& {{\mathcal{L}}_{\text{CL}}}(DEF,{{x}_{s}},{{c}^{+}},c_{1}^{-},\ldots e_{2B-1}^{-}) = \nonumber \\
& \ \ \ \ \ \ \ \ \ \ \ -\log \frac{{{e}^{\text{Sim}({{x}_{s}},{{c}^{+}})}}}{{{e}^{\text{Sim}({{x}_{s}},{{c}^{+}})}}+\sum\nolimits_{j=1}^{2B-1}{{{e}^{\text{Sim}({{x}_{s}},e_{j}^{-})}}}} \label{eq6}
\end{align}

\subsection{Task Inference}
We obtain the instruction template for inference by retrieving the top-$k$ examples using the retriever. The LM predicts the probability of generating a target $y_s$, trained by minimizing the negative likelihood loss:

\begin{align}
&{{\mathcal{L}}_{\text{NLL}}}= \nonumber \\
&-\sum\limits_{l=1}^{L}{\log }p(y_{s}^{l}\mid DEF,\{{{\hat{x}}_{i}},{{\hat{y}}_{i}}\}_{i=1}^{k},{{x}_{s}},y_{s}^{<l}) \label{eq7}
\end{align}

\noindent where $L$ is the output length of the target sample.
	
The LM receives input only in the form of instructions. Examples can be converted into instruction prompts using ATE, ATSC, and ASPE templates. To train the LM, only the top-1 scoring example is applied for fine-tuning,
\begin{align}
{{\tilde{y}}_{s}}=\text{LM}(\text{Tmpl}(DEF,\hat{x}_{1}^{+},\hat{y}_{1}^{+},{{x}_{s}})) \label{eq8}
\end{align}

For inference, instruction prompts are constructed using the top-$k$ examples. Upon receiving instructions, the LM produces an output specific to the current task and input $x_s$,
\begin{align}
{{\tilde{y}}_{s}}=\text{LM}(\text{Tmpl}(DEF,\{{{\hat{x}}_{i}},{{\hat{y}}_{i}}\}_{i=1}^{k},{{x}_{s}})) \label{eq9}
\end{align}

\subsection{Alternating Training Schema}

\begin{figure}[!t]
    \centering  
    \centerline{\includegraphics[width=0.49\textwidth]{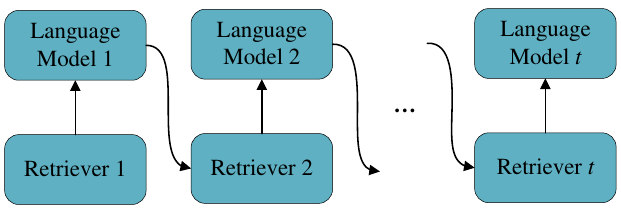}}
    \caption{Alternating training schema.}
    \label{a5}
\end{figure}
The training of a retriever depends on the scoring of the LM. However, the LM training requires the retriever to form an instruction template. The instruction tuning performance depends on the collaborative effort between the two models.

Thus, we adopted an alternating training schema for the retriever and language models, as shown in Figure \ref{a5}. In particular, the finetuned LM in the $t-1$ step is used as a scoring model to train the retriever in the $t$ step. The LM in the $t$ step is finetuned by the instruction generated by the updated retriever.

\section{Experiments}

\begin{table}[!t]
\resizebox{\linewidth}{!}{
\begin{tabular}{ccccccc}
\toprule
\textbf{Dataset}                 & \textbf{Split} & \textbf{\#Pos} & \textbf{\#Neg} & \textbf{\#Neu} & \textbf{\#No} & \textbf{\#T} \\ \midrule
\multirow{2}{*}{\textbf{Lap14}}  & train          & 987          & 866          & 460          & 1557        & 3045           \\
                                 & test           & 341          & 128          & 169          & 378         & 800            \\ \hline
\multirow{2}{*}{\textbf{Rest14}} & train          & 2164         & 805          & 633          & 1020        & 3041           \\
                                 & test           & 728          & 196          & 196          & 194         & 800            \\ \hline
\multirow{2}{*}{\textbf{Rest15}} & train          & 912          & 256          & 36           & 482         & 1315           \\
                                 & test           & 326          & 182          & 34           & 284         & 685            \\ \hline
\multirow{2}{*}{\textbf{Rest16}} & train          & 1240         & 439          & 69           & 766         & 2000           \\
                                 & test           & 468          & 117          & 30           & 256         & 676      \\
\bottomrule
\end{tabular}}
\caption{Statistics for experiment datasets. \#Pos, \#Neg, and \#Neu denote the number of aspects with positive, negative, and neutral sentiments, respectively, \#No denotes the number of aspect-free terms, and \#T denotes the total number of samples.}
\label{dataset}
\end{table}

\subsection{Datasets}

Experiments were conducted on the Semeval-2014, 15, and 16 datasets \cite{Pontiki2014,Pontiki2015,Pontiki2016}, which are benchmark dataset for the ABSA task. The benchmark comprises four datasets, including customer reviews from two domains, laptops (\textbf{Lap14}) and restaurants (\textbf{Rest14}, \textbf{Rest15}, and \textbf{Rest16}). The model performance was measured using $F_1$ scores for the ATE and ASPE tasks and accuracy for the ATSC task. Table \ref{dataset} provides the details of the data distribution. Conflict labels were ignored. 

\subsection{Implementation Details}

The LM was initialized using the \textbf{flan-t5-base}\footnote{https://huggingface.co/google/flan-t5-base}. AdamW \cite{Loshchilov2019} was applied to optimize the model with an initial learning rate of 5e-5. The training batch size was 2. The gradient accumulation steps were set to 2. The number of epochs was 4 for the retriever, and 2 for the language model. The maximum sequence length was 128. Data ratio $r$ of the training retriever was set to 0.1. The number of examples was one for the training phase and from zero to seven for the evaluation phase; in the comparative experiments, $k$ was set to 4. The step $t$ of alternating training schema was 3. 

\subsection{Baselines}

\begin{figure*}[!t]    
  \centering            
  \subfloat[ATE]   
  {
        \label{a}
        \includegraphics[width=0.33\textwidth]{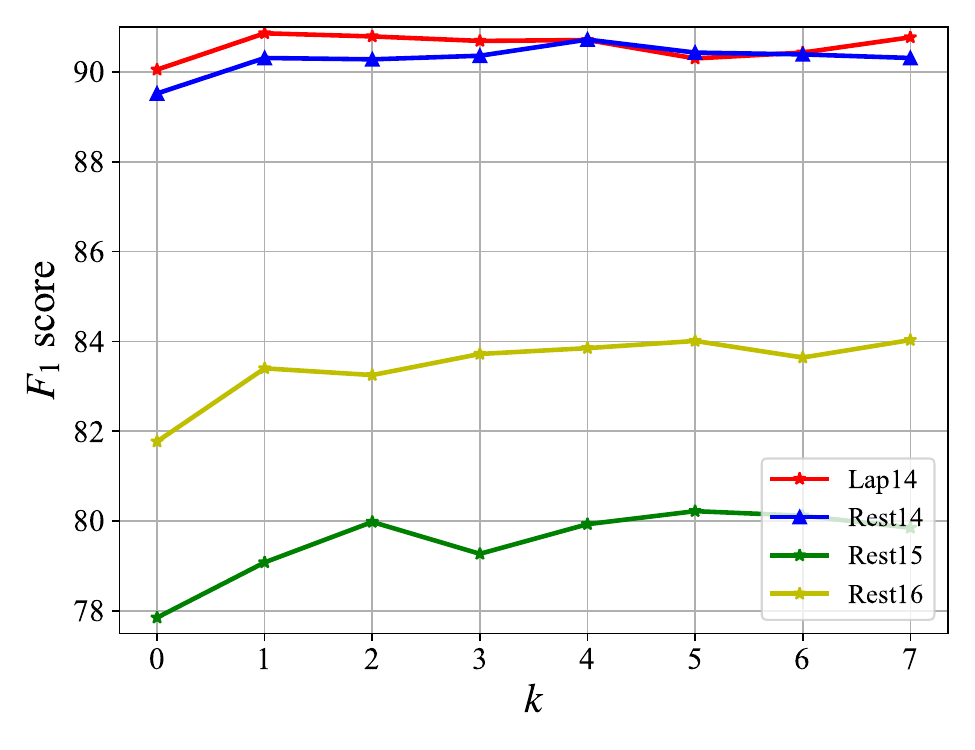}
  }
  \subfloat[ATSC]
  {
      \label{b}
        \includegraphics[width=0.33\textwidth]{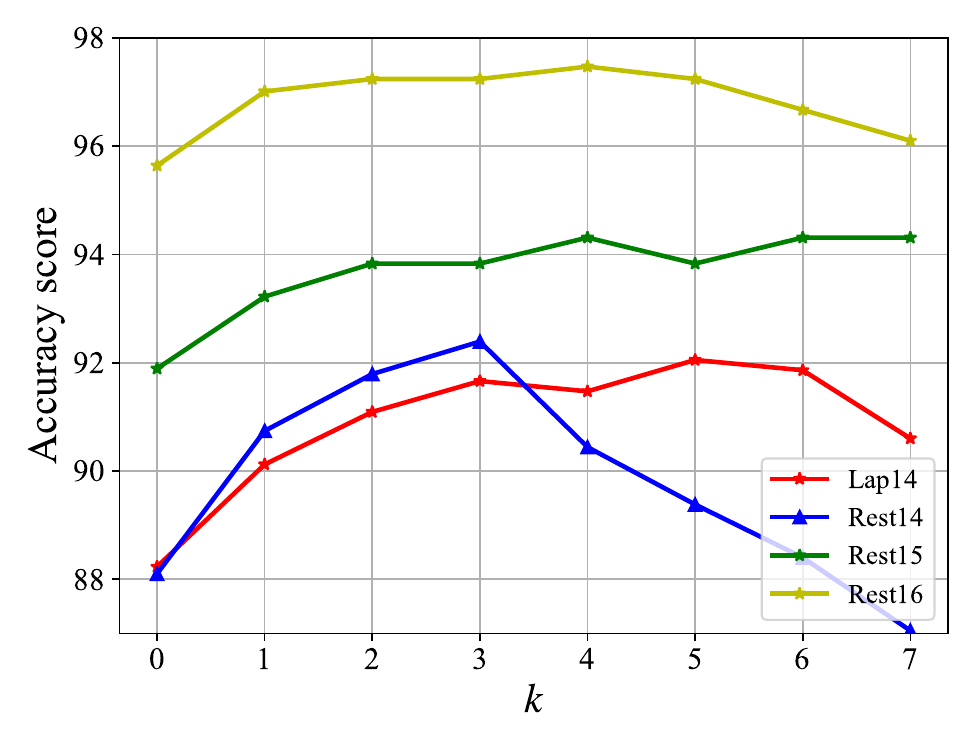}
  }
  \subfloat[ASPE]
  {
      \label{c}
        \includegraphics[width=0.33\textwidth]{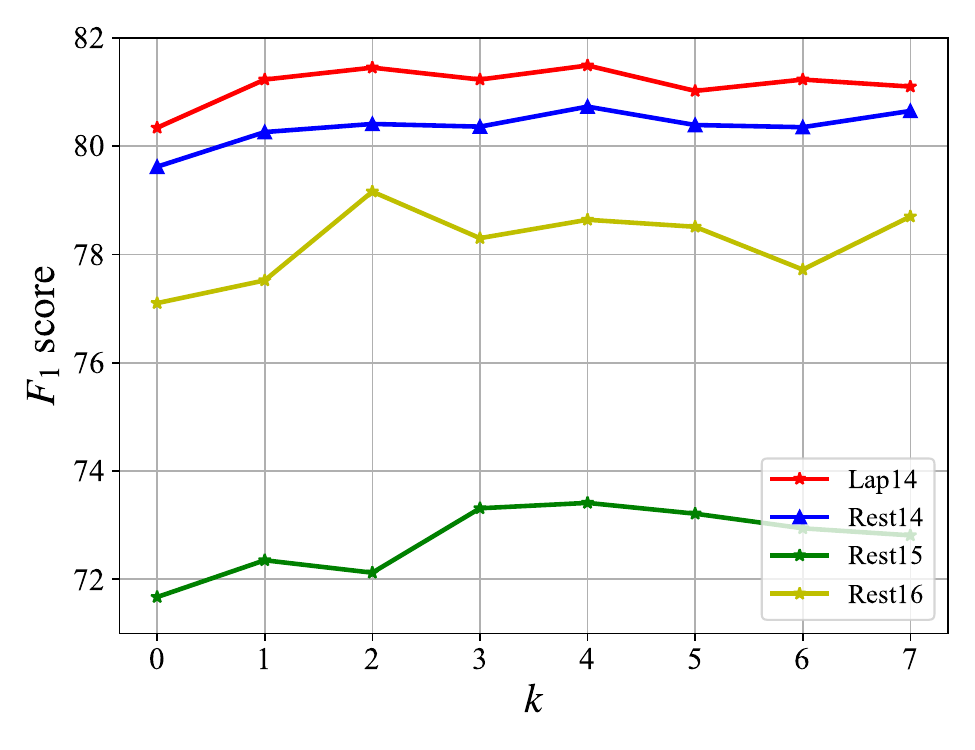}
  }
  \caption{The impact of the number of examples on performance (\%) on \textbf{Rest14}.}    
  \label{a6}            
\end{figure*}

The baseline models that emerged from the comparative experiments were categorized into generative and non-generative models,

\begin{itemize}

\item \noindent\textbf{1) Generative methods:} GPT2$_{\text{med}}$ \cite{Hosseini}, BARTABSA \cite{Yan2021}, GAS \cite{Zhang2021}, IT-MTL \cite{Varia2022}, InstructABSA \cite{Scaria2023};

\item \noindent\textbf{2) Non-generative methods:} SPAN \cite{hu-etal-2019}, GRACE \cite{Luo2020}, ABSA-DeBERTa \cite{Ricardo}, LSAT \cite{Yang2021}, RACL-BERT \cite{Chen2020}, Dual-MRC \cite{Mao2021}, Seq2Path \cite{mao-etal-2022-seq2pat}.

\end{itemize}

The number of paths $k$ in Seq2Path is 4 and the results are from \citet{mao-etal-2022-seq2pat}. The results of other baselines are from \citet{Scaria2023}.
A more detailed descriptions about baselines are provided in Appendix \ref{sec:appendix}.

\subsection{Comparative Results}

\begin{table}[!t]
\centering
\resizebox{\linewidth}{!}{
\begin{tabular}{ccccc}
\toprule
\textbf{Model} & \textbf{Lap14} & \textbf{Rest14} & \textbf{Rest15} & \textbf{Rest16} \\ \midrule
GRACE          & 87.93          & 85.45           & -               & -               \\
SPAN          & 83.35          & 82.38          & -               & -               \\
GPT2$_{\text{med}}$        & 82.04          & 75.94           & -               & -               \\
BARTABSA       & 83.52          & 87.07           & 75.48           & -               \\
IT-MTL         & 76.93          & -               & 74.03           & 79.41           \\
InstructABSA1  & 91.40          & \textbf{92.76}  & 75.23           & 81.48           \\
InstructABSA2  & \textbf{92.30} & 92.10           & 76.64           & 80.32           \\ \hline
Ours ($k=4$)     & 90.05          & 90.72           & \textbf{79.93}  & \textbf{83.85} \\ \bottomrule
\end{tabular}}
\caption{ATE subtask results in terms of the $F_1$ scores (\%).}
\label{res:ate}
\end{table}

\begin{table}[!t]
\centering
\resizebox{\linewidth}{!}{
\begin{tabular}{ccccc}
\toprule
\textbf{Model} & \textbf{Lap14} & \textbf{Rest14} & \textbf{Rest15} & \textbf{Rest16} \\ \midrule
SPAN   & 81.39          & 89.95           & -               & -               \\
ABSA-DeBERTa   & 82.76          & 89.46           & -               & -               \\
LSAT           & 86.31          & \textbf{90.86}  & -               & -               \\
RACL-BERT      & 73.91          & 81.61           & 74.91           & -               \\
Dual-MRC       & 75.97          & 82.04           & 73.59           & -               \\
InstructABSA1  & 80.62          & 86.25           & 83.02           & 89.1            \\
InstructABSA2  & 81.56          & 85.17           & 84.5            & 89.43           \\ \hline
Ours ($k=4$)     & \textbf{91.47} & 90.44           & \textbf{94.31}  & \textbf{97.47} \\ \bottomrule
\end{tabular}}
\caption{ATSC subtask results in terms of the accuracy (\%).}
\label{res:atsc}
\end{table}

Tables \ref{res:ate}, \ref{res:atsc}, and \ref{res:aspe} summarize the results of the proposed method relative to those of the baseline methods. The highest performance is in bold.

For the ATE task, the proposed method achieved $F_1$ scores of 79.93 and 83.85 on the \textbf{Rest15} and \textbf{Rest16} datasets, respectively, exceeding the baselines by 3.29\% and 2.37\%, respectively. In addition, it achieved better scores on the \textbf{Lap14} and \textbf{Rest14} datasets but was approximately 2\% lower than InstructABSA, which used fixed examples. This might be related to the instruction template and the pre-trained language model. InstructABSA utilizes examples that distinguish between sentiment polarity and employs the pre-trained language model \textbf{tk-instruct-base-def-pos}\footnote{https://huggingface.co/allenai/tk-instruct-base-def-pos}, which could contribute to the performance variation. Nevertheless, the proposed model only performs worse than InstructABSA on the \textbf{Rest14} and \textbf{Lap14} datasets for the ATE task. We speculate two potential reasons for this: 
\begin{enumerate}
    \item The \textbf{Rest14} and \textbf{Lap14} datasets are larger thus have more retrievable examples compared to \textbf{Rest15} and \textbf{Rest16}, resulting in increased uncertainties in example variations. This might lead the model to rely on the knowledge provided by the examples overly.
    \item The ATE task might be relatively straightforward and offer limited knowledge for improvement.
\end{enumerate}

For the ATSC task, the proposed method achieved optimal performances on the \textbf{Lap14}, \textbf{Rest15}, and \textbf{Rest16} datasets, significantly outperforming the baselines (5-10\% improvement) and achieving a similar performance to that of LSAT on the \textbf{Rest14} dataset.

For the ASPE task, the proposed method surpassed the baselines on all four datasets, obtaining $F_1$ scores of 81.49, 80.73, 73.41, and 78.64.

Moreover, for the ATSC task, the model performed better on smaller datasets (\textbf{Rest15}, \textbf{Rest16}) than on larger datasets (\textbf{Lap14} and \textbf{Rest14}).

\begin{table}[!t]
\centering
\resizebox{\linewidth}{!}{
\begin{tabular}{ccccc}
\toprule
\textbf{Model} & \textbf{Lap14} & \textbf{Rest14} & \textbf{Rest15} & \textbf{Rest16} \\ \midrule
GRACE          & 75.97          & 78.07           & -               & -               \\
SPAN            & 68.06          & 74.92           & -               & -               \\
GPT2$_{\text{med}}$        & 53.55          & 60.07           & -               & -               \\
GAS            & 68.64          & 77.13           & 66.78           & 73.64           \\
Seq2Path       & 70.00          & 77.01           & 68.35           & 75.87           \\
BARTABSA       & 67.37          & 73.56           & 66.61           & -               \\
IT-MTL         & 66.07          & -               & 67.06           & 74.07           \\
InstructABSA1  & 78.89          & 76.16           & 69.02           & 74.24           \\
InstructABSA2  & 79.34          & 79.47           & 69.39           & 73.06           \\ \hline
Ours ($k=4$)     & \textbf{81.49} & \textbf{80.73}  & \textbf{73.41}  & \textbf{78.64} \\ \bottomrule
\end{tabular}}
\caption{ASPE subtask results in terms of the $F_1$ scores (\%).}
\label{res:aspe}
\end{table}

\begin{table*}[!t]
\resizebox{\linewidth}{!}{
\begin{tabular}{lcccccccccccc}
\toprule
\multicolumn{1}{c}{\multirow{2}{*}{\textbf{Model}}} & \multicolumn{4}{c}{\textbf{ATE ($F_1$)}}                     & \multicolumn{4}{c}{\textbf{ATSC ($Acc$)}}                   & \multicolumn{4}{c}{\textbf{ASPE ($F_1$)}}                    \\
\multicolumn{1}{c}{}                                & \textbf{L14} & \textbf{R14} & \textbf{R15} & \textbf{R16} & \textbf{L14} & \textbf{R14} & \textbf{R15} & \textbf{R16} & \textbf{L14} & \textbf{R14} & \textbf{R15} & \textbf{R16} \\ \midrule
Ours ($k=4$)                      & 90.05        & 90.72        & 79.93        & 83.85        & 91.47        & 90.44        & 94.31        & 97.47        & 81.49        & 80.73        & 73.41        & 78.64        \\
\,\,\,\,\,\,w/o alternating training                            & 89.82        & 89.13        & 77.46        & 80.16        & 91.37        & 89.47        & 93.46        & 96.55        & 78.05        & 77.63        & 69.05        & 77.61        \\
\,\,\,\,\,\,w/o retriever                                       & 89.77        & 88.34        & 75.59        & 80.06        & 87.11        & 87.87        & 91.64        & 96.09        & 79.16        & 78.11        & 70.78        & 77.55        \\
\,\,\,\,\,\,w/o example                                         & 89.74        & 88.03        & 76.00        & 79.79        & 87.59        & 87.95        & 91.52        & 95.86        & 77.05        & 76.20        & 68.29        & 76.02        \\
\,\,\,\,\,\,w/o instruction                                     & 88.68        & 87.56        & 74.08        & 79.58        & 87.30        & 86.60        & 90.67        & 95.75        & 76.66        & 74.97        & 68.62        & 76.01      \\
\bottomrule
\end{tabular}
}
\caption{Ablation study (\%). L14, R14, R15, and R16 denote the datasets \textbf{Lap14}, \textbf{Rest14}, \textbf{Rest15}, and \textbf{Rest16}, respectively.}
\label{ablation}
\end{table*}

\begin{table*}[t!]
\resizebox{\linewidth}{!}{
\begin{tabular}{ccccccccccccc}
\toprule
\multicolumn{1}{c}{\multirow{2}{*}{\textbf{Model}}} & \multicolumn{4}{c}{\textbf{ATE ($F_1$)}} & \multicolumn{4}{c}{\textbf{ATSC ($Acc$)}} & \multicolumn{4}{c}{\textbf{ASPE ($F_1$)}} \\
\multicolumn{1}{c}{}                                & \textbf{L14}     & \textbf{R14}     & \textbf{R15}     & \textbf{R16}     & \textbf{L14}      & \textbf{R14}      & \textbf{R15}     & \textbf{R16}     & \textbf{L14}      & \textbf{R14}     & \textbf{R15}     & \textbf{R16}     \\ \midrule
Frozen LM ($k=4$)   & 41.96   & 25.85   & 40.77   & 39.00   & 48.83    & 65.74    & 55.21   & 62.92   & 29.04    & 12.46   & 29.83   & 25.15   \\
Ours ($k=4$)                                          & 90.05   & 90.72   & 79.93   & 83.85   & 91.47    & 90.44    & 94.31   & 97.47   & 81.49    & 80.73   & 73.41   & 78.64   \\
$\uparrow$ (\%)  & 46.60 & 28.49 & 51.01 & 46.51 & 53.38  & 72.69  & 58.54 & 64.55 & 35.64  & 15.43 & 40.63 & 31.98 \\ \bottomrule
\end{tabular}
}
\caption{Exploring the improvement of fine-tuning language models. \textbf{Frozen LM} indicates that the parameters of LM are not updated during the training phase, which is similar to the retrieval methods used on LLM. $\uparrow$ (\%) represents how much performance improvement ratio our method has compared to \textbf{Frozen LM}.}
\label{aaa}
\end{table*}

\subsection{Number of Examples}

To explore the impact of the number of examples on the performance, we used a retriever to extract different numbers of examples for model inference. The maximum number of examples was 7 to prevent the input text from exceeding the maximum sequence length of the model. Figure \ref{a6} shows the results.

Overall, the performances show two trends as the number of examples increases: (i) rising and then falling or (ii) rising. The use of examples allows the model to achieve a significant performance gain. As the number of examples increases, the model can acquire more knowledge from beneficial examples. However, this performance enhancement trend is not always significant, and an increase in the number of examples can harm the model performance. Owing to the limited capacity of the retriever and limited number of beneficial examples for the query in the example pool, not all retrieved examples were beneficial or harmful to the query.

\subsection{Ablation Study}

Table \ref{ablation} presents the results of the ablation study used to investigate the effectiveness of each component; \textbf{w/o alternating training} means that alternating training schema was not used; \textbf{w/o retriever}, the retriever was removed and fixed examples were used; \textbf{w/o example}, examples were not used; and \textbf{w/o instruction}, instruction prompts were not used. The results demonstrate the effectiveness of each part of the proposed method.
 
For further analysis, the performance decrease in the term \textbf{w/o alternating training} suggests that alternating training schema can better narrow the gap between the retriever and LM. Additionally, because the term \textbf{w/o retriever} outperforms the term \textbf{w/o example} overall, although inferior to the term \textbf{w/o example} in some cases, suggests that the model can learn from the examples, but fixed examples function differently for different target samples and are not conducive to inference for some extreme situations.

\subsection{The Role of Fine-tuning Language Models}
The Table \ref{aaa} presents the results of our experiments.  For the ATSC task, phrases such as \textit{The pizza is good} and \textit{I think the pizza was good} do not require the LM to focus on semantic similarity or the connection between aspects and opinions. The LM can infer the result based on the retrieved example sentence's \textit{output: positive}, resulting in decent performance even without training. For the ATE task, the performance is poorer due to the low overlap in output results. For the more complex ASPE task, the retrieved example sentences should provide diverse assistance to the test sentences (including structural assistance and overlapping aspects). However, an untrained LM lacks this ability and tends to search for answers from the context rather than making inferences.

We further explored the retrieval results of the method that only trains the retriever. Compared to the proposed method, it favors retrieving example sentences with overlapping results and rarely retrieving other types of examples mentioned in the Case Study section. This limitation may hinder its effectiveness in handling complex tasks that require a deeper understanding of the context and relationships between different aspects.

\begin{table}[!t]
\centering
\resizebox{\linewidth}{!}{
\begin{tabular}{ccccc}
\toprule
\textbf{Model} & \textbf{Size} & \textbf{ATE} & \textbf{ASTC} & \textbf{ASPE} \\ \midrule
T5-base        & 223M          & 89.91        & 87.34         & 79.12         \\
T5-large       & 738M          & 91.71        & 91.04         & 81.23         \\ \hline
Flan-T5-base   & 248M          & 90.72        & 90.44         & 80.73         \\
Flan-T5-large  & 783M          & 92.64        & 92.97         & 82.61 \\ \bottomrule    
\end{tabular}}
\caption{Performance (\%) with different language models on \textbf{Rest14}.}
\label{backbone}
\end{table}

\subsection{Effectiveness on Different Language Models}

The efficacy of our proposed method was examined across various backbone language models. The performance metrics for different variants in terms of sizes and types are detailed in Table \ref{backbone}. The results underscore the robust effectiveness of our proposed method across commonly used backbone models. Notably, employing larger models leads to discernible performance improvements. Particularly, the instruction pre-training variant, flan-t5, exhibits more substantial gains. This can be attributed to the smaller gap between the pre-training and fine-tuning phases.

\subsection{Case Study}

Table \ref{case} provides an overview of the retrieval results for specific queries in the ASPE task. For clarity, only the first result retrieved is presented for each query. Notably, examples with IDs 1, 3, and 4 exhibit a structural similarity to the query. For instance, ID 2 mirrors the query's aspects with the same sentiment polarity. While these instances showcase evidently favorable retrieval outcomes, there are cases where the utility of results is unclear. For instance, example ID 5 does not seem directly related to the query; however, the prediction for that particular query is accurate. Determining whether this is due to the LM disregarding irrelevant examples or if the example somehow contributes to the LM's understanding remains an open question.

\begin{table}[!t]
\resizebox{\linewidth}{!}{
\begin{tabular}{p{0.05in}p{1.1in}p{1.68in}}
\toprule                         
\textbf{ID} & \multicolumn{1}{c}{\textbf{Query}}                      & \multicolumn{1}{c}{\textbf{Example}}                                                                                                          \\ \midrule
1           & Green Tea creme brulee is a must!                       & \textbf{input:} (The asparagus, truffle oil, parmesan bruschetta is a winner!) \newline\textbf{output:}  asparagus, truffle oil, parmesan bruschetta: positive \\
2           & Serves really good sushi.                               & \textbf{input:} Best. Sushi. Ever. \newline\textbf{output:}  Sushi: positive                                                                                   \\
3           & AMAZING.                                                & \textbf{input:} Unbelievable. \newline\textbf{output:}  noaspectterm: none                                                                                     \\
4           & The food was almost always EXCELLENT.                   & \textbf{input:} The food was good. \newline\textbf{output:}  food: positive                                                                                    \\
5           & I never had an orange donut before so I gave it a shot. & \textbf{input:} this one is definintely my least favorite. \newline\textbf{output:}  noaspectterm: none              \\
\bottomrule
                                       
\end{tabular}}
\caption{Best retrieval results in the ASPE task.}
\label{case}
\end{table}

\section{Conclusion}
In this study, we proposed a retrieval-based example mining method for instructional learning in ABSA tasks to improve the performance by selecting effective examples. The proposed method conducts the alternating training of a retriever and LM by employing a two-stage training framework and iterative evolution training scheme. Experiments validated its effectiveness across ATE, ATSC, and ASPE tasks, outperforming existing baseline models. 

Future work will extend the proposed method to other tasks and models and refine the training strategies to achieve further performance gains.

\section*{Limitations}
There are three main limitations to our work compared to previous efforts:
\begin{enumerate}
    \item The mechanism of retrieval is based on the likelihood score given by the language model, however, this score only focuses on the improvement of the model's performance, and the syntactic mechanism of its work remains to be explored.
    \item The choice of the number k of examples constrains the performance of the model. Although we explored the impact of using different numbers of examples on the overall performance, it is undeniable that the optimal number of examples varies for different test inputs. Model performance would be further improved if the appropriate number of examples could be customized for each input.
    \item The proposed method is only experimented on the English dataset. Whether it works equally well and whether the retrieved examples have commonalities is still unknown in other languages (e.g., Russian, French, Chinese, etc.). It remains to be explored whether the method will work on mixed language and multilingual datasets.
\end{enumerate}

\section*{Acknowledgements}

This work was supported by the National Natural Science Foundation of China (NSFC) under Grant Nos.61966038 and 62266051, the Ministry of Science and Technology, Taiwan, ROC, under Grant No.MOST 111-2628-E-155-001-MY2, and the Exam-Exempted Postgraduate Research and Innovation Foundation of Yunnan University under Grant No.TM-23237123. The authors would like to thank the anonymous reviewers for their constructive comments.

\bibliography{anthology,custom}
\bibliographystyle{acl_natbib}

\appendix

\section{Baseline Models}
\label{sec:appendix}

The baseline models that emerged from the comparative experiments were categorized into generative and non-generative models. This section describes these baseline models in detail.

\noindent\textbf{1) Generative methods:} 
\begin{itemize}
    \item \textbf{GPT2$_{\text{med}}$} \cite{Hosseini} utilizes unidirectional self-attention and language modeling loss to capture contextual representations and leverage supervision during training. 
    \item \textbf{BARTABSA} \cite{Yan2021} formulates the ABSA extraction and classification tasks as a unified index generation problem. 
    \item \textbf{GAS} \cite{Zhang2021} formulates each task as a generative problem and predictive normalization strategy to optimize the generated outputs.
    \item \textbf{IT-MTL} \cite{Varia2022} treats ABSA as a sequence-to-sequence modeling task based on instruction tuning, achieving excellent performance with a few shots.
    \item \textbf{InstructABSA} \cite{Scaria2023} constructs fixed instruction prompts for different tasks to train the Tk-instruct model \cite{Wang2022}, and the examples in the prompts are obtained from combinations of sentiment-positive, -negative, and -neutral examples. InstructABSA1 includes two positive and two negative sentiment examples, while InstructABSA2 adds two neutral sentiment examples. 
\end{itemize}

\noindent\textbf{2) Non-generative methods:} 
\begin{itemize}
    \item \textbf{SPAN} \cite{hu-etal-2019} use a span-based labeling scheme to find and classify opinion targets in a sentence, which mitigates the problem of sentimental inconsistencies at the span level.
    \item \textbf{GRACE} \cite{Luo2020} employs cascade labeling to enhance the interaction between aspect terms and mitigates the labeling imbalance through a gradient harmonization approach.  
    \item \textbf{ABSA-DeBERTa} \cite{Ricardo} uses a decoupled attention mechanism to separate location and content vectors for sentiment analysis. 
    \item \textbf{LSAT} \cite{Yang2021} introduces a local sentiment aggregation paradigm that facilitates fine-grained sentiment consistency modeling.  
    \item \textbf{RACL-BERT} \cite{Chen2020} allows subtasks to work together in stacked multilayer networks via multitask learning and relationship propagation mechanisms. 
    \item \textbf{Dual-MRC} \cite{Mao2021} converts the original triplet extraction task into two machine reading comprehension (MRC) problems, and jointly trains multiple subtasks.
    \item \textbf{Seq2Path} \cite{mao-etal-2022-seq2pat} transforms the generation order of sentiment tuples into tree paths. This approach not only effectively addresses the issue of one aspect entity corresponding to multiple opinion words, but also ensures that the generation of each path is independent.
\end{itemize}

\end{document}